\pdfoutput=1 %

\documentclass[11pt,a4paper]{article}
\usepackage[hyperref]{acl2020}
\usepackage{times}
\usepackage{latexsym}

\usepackage{microtype}

\usepackage[utf8]{inputenc}
\usepackage[T1]{fontenc} %
\usepackage{todonotes} %
\usepackage{multirow}
\usepackage{booktabs}
\usepackage{amsmath}
\usepackage{amsfonts}

\usepackage{seqsplit}

\aclfinalcopy %

\title{A Call for More Rigor in Unsupervised Cross-lingual Learning}

\author{Mikel Artetxe$^{\dag}$\thanks{\hspace{1.5 mm}Equal contribution.}~, \ Sebastian Ruder$^{\ddag}$\footnotemark[1]~, \ Dani Yogatama$^{\ddag}$, \ Gorka Labaka$^{\dag}$, \ Eneko Agirre$^{\dag}$ \\
$^{\dag}$HiTZ Center, University of the Basque Country (UPV/EHU)\\
$^{\ddag}$DeepMind\\
\texttt{\{mikel.artetxe,gorka.labaka,e.agirre\}@ehu.eus} \\ 
\texttt{\{ruder,dyogatama\}@google.com}
}

\date{}

\begin{document}
\maketitle
\begin{abstract}
We review motivations, definition, approaches, and methodology for unsupervised cross-lingual learning and call for a more rigorous position in each of them.
An existing rationale for such research is based
on the lack of parallel data for many of the world's languages.
However, we argue that a scenario without \emph{any} parallel data and abundant monolingual data is unrealistic in practice.
We also discuss different training signals
that have been used in previous work,
which depart from the pure unsupervised setting.
We then describe common methodological 
issues in tuning and evaluation of unsupervised cross-lingual models
and present best practices.
Finally, we provide a unified outlook for 
different types of research in
this area (i.e., cross-lingual word embeddings,
deep multilingual pretraining, and unsupervised machine translation)
and argue for comparable evaluation of these models.

\end{abstract}

\section{Introduction}

The study of the connection among human languages has contributed to major discoveries including the evolution of languages, the reconstruction of proto-languages, and an understanding of language universals \cite{eco1995search}.
In natural language processing, the main promise of multilingual learning is to bridge the digital language divide, to enable access to information and technology for the world's 6,900 languages \cite{Ruder2019survey}. For the purpose of this paper,
 we define \textit{``multilingual learning''} as learning a common model for two or more languages from raw text, without any downstream task labels. Common use cases include translation as well as pretraining multilingual representations.
We will use the term interchangeably with \textit{``cross-lingual learning''}.

Recent work in this direction has increasingly focused on purely unsupervised cross-lingual learning (UCL)---i.e., cross-lingual learning without any parallel signal across the languages. We provide an overview in \S\ref{sec:background}. Such work has been motivated by the apparent dearth of parallel data for most of the world's languages. In particular, previous work has noted that \textit{``data encoding cross-lingual equivalence is often expensive to obtain''} \citep{zhang2017adversarial} whereas \textit{``monolingual data is much easier to find''} \citep{lample2018unsupervised}. Overall, it has been argued that unsupervised cross-lingual learning \textit{``opens up opportunities for the processing of extremely low-resource languages and domains that lack parallel data completely''} \citep{zhang2017adversarial}.

We challenge this narrative and argue that the scenario of no parallel data \emph{and} sufficient monolingual data is unrealistic and not reflected in the real world (\S\ref{subsec:practical}). Nevertheless, UCL is an important research direction and we advocate for its study based on an inherent scientific interest (to better understand and make progress on general language understanding), usefulness as a lab setting, and simplicity (\S\ref{subsec:motivation}). 

\emph{Unsupervised} cross-lingual learning permits no supervisory signal
by definition. However, previous work implicitly includes monolingual and cross-lingual signals that constitute a departure from the pure setting. We review existing training signals as well as other signals that may be of interest for future study (\S\ref{sec:unsupervised_signals}).
We then discuss methodological issues in UCL (e.g., validation, hyperparameter tuning) and propose best evaluation practices
(\S\ref{sec:methodological_issues}). 
Finally, we provide a unified outlook of established research areas 
(cross-lingual word embeddings, deep multilingual models and unsupervised machine translation) in UCL (\S\ref{sec:unsupervised_types}), and conclude with a summary of our recommendations (\S\ref{sec:recommendations}).

\section{Background} \label{sec:background}

In this section, we briefly review existing work on UCL, covering cross-lingual word embeddings (\S\ref{subsec:clwe}), deep multilingual pre-training (\S\ref{subsec:mbert}), and unsupervised machine translation (\S\ref{subsec:umt}).

\subsection{Cross-lingual word embeddings} \label{subsec:clwe}

Cross-lingual word embedding methods traditionally relied on parallel corpora \citep{gouws2015bilbowa,luong2015bilingual}. Nonetheless, the amount of supervision required was greatly reduced via cross-lingual word embedding mappings, which work by separately learning monolingual word embeddings in each language and mapping them into a shared space through a linear transformation. Early work required a bilingual dictionary to learn such a transformation \citep{mikolov2013exploiting,faruqui2014improving}. 
This requirement was later reduced with self-learning \citep{artetxe2017learning}, and ultimately removed via unsupervised initialization heuristics \citep{artetxe2018robust,hoshen2018nonadversarial} and adversarial learning \citep{zhang2017adversarial,conneau2018word}.
Finally, several recent methods have formulated cross-lingual embedding alignment as an optimal transport problem \citep{zhang2017earth,grave2018unsupervised,alvarezmelis2018gromov}.

\subsection{Deep multilingual pretraining} \label{subsec:mbert}

Following the success in learning shallow word embeddings \citep{mikolov2013distributed,pennington2014glove}, there has been an increasing interest in learning contextual word representations
\citep{dai2015semisupervised,peters2018deep,howard2018universal}. Recent research has been dominated by BERT \citep{devlin2018bert}, which uses a bidirectional transformer encoder trained on masked language modeling and next sentence prediction, which led to impressive gains on various downstream tasks.

While the above approaches are limited to a single language, a multilingual extension of BERT (mBERT) has been shown to also be effective at learning cross-lingual representations in an unsupervised way.\footnote{\url{https://github.com/google-research/bert/blob/master/multilingual.md}} The main idea is to combine monolingual corpora in different languages, upsampling those with less data, and training a regular BERT model on the combined data. \citet{conneau2019crosslingual} follow a similar approach but perform a more thorough evaluation and report substantially stronger results,\footnote{The full version of their model (XLM) requires parallel corpora for their translation language modeling objective, but the authors also explore an unsupervised variant using masked language modeling alone.} which was further scaled up by \citet{conneau2019unsupervised}. Several recent studies \cite{Wu2019,Pires2019,Artetxe2019crosslingualtransfer,wu2019emerging} analyze mBERT
to get a better understanding of its capabilities.

\subsection{Unsupervised machine translation} \label{subsec:umt}

Early attempts to build machine translation systems using monolingual data alone go back to statistical decipherment \citep{ravi2011deciphering,dou2012large,dou2013dependency}. However, this approach was only shown to work in limited settings, and the first convincing results on standard benchmarks were achieved by \citet{artetxe2018nmt} and \citet{lample2018unsupervised} on unsupervised Neural Machine Translation (NMT). Both approaches rely on cross-lingual word embeddings to initialize a shared encoder, and train it in conjunction with the decoder using a combination of denoising autoencoding, back-translation, and optionally adversarial learning.

Subsequent work adapted these principles to unsupervised phrase-based Statistical Machine Translation (SMT), obtaining large improvements over the original NMT-based systems \citep{lample2018phrase,artetxe2018smt}. This alternative approach uses cross-lingual $n$-gram embeddings to build an initial phrase table, which is combined with an n-gram language model and a distortion model, and further refined through iterative back-translation. 
There have been several follow-up attempts to combine NMT and SMT based approaches \citep{marie2018unsupervised,ren2019unsupervised,artetxe2019effective}. 
More recently, \citet{conneau2019crosslingual}, \citet{Song2019mass} and \citet{liu2020multilingual} obtain strong results using deep multilingual pretraining rather than cross-lingual word embeddings to initialize unsupervised NMT systems.

\section{Motivating fully unsupervised learning} \label{sec:motivation}

In this section, we challenge the narrative of motivating UCL based on a lack of parallel resources. We argue that the strict unsupervised scenario cannot be motivated from an immediate practical perspective, and elucidate what we believe should be the true goals of this research direction.

\subsection{How practical is the strict unsupervised scenario?} \label{subsec:practical}

Monolingual resources subsume parallel resources. For instance, each side of a parallel corpus effectively serves as a monolingual corpus.
From this argument, it follows that monolingual data is cheaper to obtain than parallel data, so unsupervised cross-lingual learning should in principle be more generally applicable than supervised learning.

However, we argue that the common claim that the requirement for \textbf{parallel data} \textit{``may not be met for many language pairs in the real world''} \citep{xu2018unsupervised} is largely inaccurate. For instance, the JW300 parallel corpus covers 343 languages with around 100,000 parallel sentences per language pair on average \citep{agic2019jw300}, and the multilingual Bible corpus collected by \citet{mayer2014creating} covers 837 language varieties (each with a unique ISO 639-3 code). Moreover, the PanLex project aims to collect multilingual lexica for all human languages in the world, and already covers 6,854 language varieties with at least 20 lexemes, 2,364 with at least 200 lexemes, and 369 with at least 2,000 lexemes \cite{kamholz2014panlex}. While 20 or 200 lexemes might seem insufficient, weakly supervised cross-lingual word embedding methods already proved effective with as little as 25 word pairs \citep{artetxe2017learning}. More recent methods have focused on completely removing this weak supervision \citep{conneau2018word,artetxe2018robust}, which can hardly be justified from a practical perspective given the existence of such resources and additional training signals stemming from a (partially) shared script (\S\ref{sec:cross-lingual_training_signals}). Finally, given the availability of sufficient monolingual data, noisy parallel data can often be obtained by mining bitext \cite{Schwenk2019wikimatrix, Schwenk2019ccmatrix}. 

In addition, large \textbf{monolingual data} is difficult to obtain for low-resource languages. For instance, recent work on cross-lingual word embeddings has mostly used Wikipedia as its source for monolingual corpora \citep{gouws2015bilbowa,Vulic2016seedlexicon,conneau2018word}. However, as of November 2019, Wikipedia exists in \textit{only} 307 languages\footnote{\url{https://en.wikipedia.org/wiki/List_of_Wikipedias}} of which nearly half have less than 10,000 articles. While one could hope to overcome this by taking the entire web as a corpus, as facilitated by Common Crawl\footnote{\url{https://commoncrawl.org/}} and similar initiatives, this is not always feasible for low-resource languages. First, the presence of less resourced languages on the web is very limited, with only a few hundred languages recognized as being used in websites.\footnote{\url{https://w3techs.com/technologies/overview/content_language}} This situation is further complicated by the limited coverage of existing tools such as language detectors \cite{buck2014ngrams,Grave2018}, which only cover a few hundred languages. Alternatively, speech could also serve as a source of monolingual data (e.g., by recording public radio stations). However, this is an unexplored direction within UCL, and collecting, processing and effectively capitalizing on speech data is far from trivial, particularly for low-resource languages.

All in all, we conclude that the alleged scenario involving no parallel data \emph{and} sufficient monolingual data is not met in the real world in the terms explored by recent UCL research. Needless to say, effectively exploiting unlabeled data is important in any low-resource setting. However, refusing to use an informative training signal---which parallel data is---when it does indeed exist, cannot be justified from a practical perspective if one's goal is to build the strongest possible model. For this reason, we believe that semi-supervised learning is a more suitable paradigm for truly low-resource languages, and UCL should not be motivated from an immediate practical perspective.

\subsection{A scientific motivation} \label{subsec:motivation}

Despite not being an entirely realistic setup, we believe that UCL is an important research direction for the reasons we discuss below.

\paragraph{Inherent scientific interest.} The extent to which two languages can be aligned based on independent samples---without any cross-lingual signal---is an open and scientifically relevant problem \textit{per se}. In fact, it is not entirely obvious that UCL should be possible at all, as humans would certainly struggle to align two unknown languages without any grounding. Exploring the limits of UCL could help to understand the limits of the principles that the corresponding methods are based on, such as the distributional hypothesis. Moreover, this research line could bring new insights into the properties and inner workings of both language acquisition and the underlying computational models that ultimately make UCL possible. Finally, such methods may be useful in areas where supervision is impossible to obtain, such as when dealing with unknown or even non-human languages. 

\paragraph{Useful as a lab setting.} The strict unsupervised scenario, although not practical, allows us to isolate and better study the use of monolingual corpora for cross-lingual learning. We believe lessons learned in this setting can be useful in the more practical semi-supervised scenario. In a similar vein, monolingual language models, although hardly useful on their own, have contributed to large improvements in other tasks. From a research methodology perspective, unsupervised systems also set a competitive baseline, which any semi-supervised method should improve upon.

\paragraph{Simplicity as a value.}
As we discussed previously, refusing to use an informative training signal when it does exist can hardly be beneficial, so we should not expect UCL to perform better than semi-supervised learning. However, simplicity is a value in its own right. Unsupervised approaches could be preferable to their semi-supervised counterparts if the performance gap between them is small enough. For instance, unsupervised cross-lingual embedding methods have been reported to be competitive with their semi-supervised counterparts in certain settings \cite{Glavas2019}, while being easier to use in the sense that they do not require a bilingual dictionary.

\section{What does \textit{unsupervised} mean?} \label{sec:unsupervised_signals}

In its most general sense, unsupervised cross-lingual learning can be seen as referring to any method relying \emph{exclusively} on monolingual text data in two or more languages. However, there are different training signals---stemming from common assumptions and varying amounts of linguistic knowledge---that one can potentially exploit under such a regime. This has led to an inconsistent use of this term in the literature. In this section, we categorize different training signals available both from a monolingual and a cross-lingual perspective and discuss additional scenarios enabled by multiple languages.

\subsection{Monolingual training signals}

From a computational perspective, text is modeled as a sequence of discrete symbols. In UCL, the training data consists of a set of such sequences in each of the languages. In principle, without any knowledge about the languages, one would have no prior information of the nature of such sequences or the possible relations between them. In practice, however, sets of sequences are assumed to be independent, and existing work differs whether they assume document-level sequences \citep{conneau2019crosslingual} or sentence-level sequences \cite{artetxe2018nmt,lample2018unsupervised}.

\paragraph{Nature of atomic symbols.} A more important consideration is the nature of the atomic symbols in such sequences. To the best of our knowledge, previous work assumes some form of word segmentation or tokenization (e.g., splitting by whitespaces or punctuation marks). Early work on cross-lingual word embeddings considered such tokens as atomic units. However, more recent work \cite{hoshen2018nonadversarial,Glavas2019} has primarily used fastText embeddings \cite{Bojanowski2017} which incorporate subword information into the embedding learning, although the vocabulary is still defined at the token level. In addition, there have also been approaches that incorporate character-level information into the alignment learning itself \cite{Heyman2017,Riley2018}. In contrast, most work on contextual word embeddings and unsupervised machine translation operates with a subword vocabulary \cite{devlin2018bert,conneau2019crosslingual}.

While the above distinction might seem irrelevant from a practical perspective, we think that it is important from a more fundamental point of view (e.g. in relation to the distributional hypothesis as discussed in \S\ref{subsec:motivation}). Moreover, some of the underlying assumptions might not generalize to different writing systems (e.g. logographic instead of alphabetic). For instance, subword tokenization has been shown to perform poorly on reduplicated words \cite{Vania2017}. In relation to that, one could also consider the text in each language as a stream of discrete character-like symbols without any notion of tokenization. Such a \emph{tabula rasa} approach is potentially applicable to any arbitrary language, even when its writing system is not known, but has so far only been explored for a limited number of languages in a monolingual setting \cite{Hahn2019tabularasa}.

\paragraph{Linguistic information.} Finally, one can exploit additional linguistic knowledge through linguistic analysis such as lemmatization, part-of-speech tagging, or syntactic parsing. For instance, before the advent of unsupervised NMT, statistical decipherment was already shown to benefit from incorporating syntactic dependency relations \citep{dou2013dependency}. For other tasks such as unsupervised POS tagging \cite{Snyder2008postagging}, monolingual tag dictionaries have been used. While such approaches could still be considered unsupervised from a cross-lingual perspective, we argue that the interest of this research direction is greatly limited by two factors: (i) from a theoretical perspective, it assumes some fundamental knowledge that is not directly inferred from the raw monolingual corpora; and (ii) from a more practical perspective, it is not reasonable to assume that such resources are available in the less resourced settings where this research direction has more potential for impact.

\subsection{Cross-lingual training signals} \label{sec:cross-lingual_training_signals}

Pure UCL should not use any cross-lingual signal by definition. When we view text as a sequence of discrete atomic symbols (either characters or tokens), a strict interpretation of this principle would consider the set of atomic symbols in different languages to be disjoint, without prior knowledge of the relationship between them.

Needless to say, any form of learning requires making assumptions, as one needs some criterion to prefer one mapping over another. In the case of UCL, such assumptions stem from the structural similarity across languages (e.g. semantically equivalent words in different languages are assumed to occur in similar contexts). In practice, these assumptions weaken as the distribution of the datasets diverges, and some UCL models have been reported to break under a domain shift \cite{Sogaard2018,guzman2019flores,marchisio2020unsupervised}. Similarly, approaches that leverage linguistic features such as syntactic dependencies may assume that these are similar across languages.

In addition, one can also assume that the 
sets of symbols that are used to represent 
different languages have some commonalities.
This departs from the strict definition of UCL above, establishing some prior connections between the sets of symbols in different languages.
Such an assumption is reasonable from a practical perspective, 
as there are a few scripts (e.g. Latin, Arabic or Cyrillic) that cover a large fraction of languages. Moreover, even when two languages use different writing systems or scripts, there are often certain elements that are still shared (e.g. Arabic numerals, named entities written in a foreign script, URLs, certain punctuation marks, etc.). In relation to that, several models have relied on identically spelled words \citep{artetxe2017learning,Smith2017,Sogaard2018} or string-level similarity across languages \citep{Riley2018,artetxe2019effective} as training signals.
Other methods use a 
joint subword vocabulary for all languages, 
indirectly exploiting the commonalities in their writing system \citep{lample2018phrase,conneau2019crosslingual}.

However, past work greatly differs on the nature and relevance that is attributed to such a training signal. The reliance on identically spelled words has been considered as a weak form of supervision in the cross-lingual word embedding literature \cite{Sogaard2018,Ruder2018discriminative}, and significant effort has been put into developing strictly unsupervised methods that do not rely on such signal \cite{conneau2018word}. In contrast, the unsupervised machine translation literature has not payed much attention to this factor, and has often relied on identical words \citep{artetxe2018nmt}, string-level similarity \citep{artetxe2019effective}, or a joint subword vocabulary \citep{lample2018phrase,conneau2019crosslingual} under the unsupervised umbrella. The same is true for unsupervised deep multilingual pretraining, where a shared subword vocabulary has been a common component \cite{Pires2019,conneau2019crosslingual}, although recent work shows that it is not important to share vocabulary across languages \cite{Artetxe2019crosslingualtransfer,wu2019emerging}.

Our position is that making assumptions on linguistics universals is acceptable and ultimately necessary for UCL. However, we believe that any connection stemming from a (partly) shared writing system belongs to a different category, and should be considered a separate cross-lingual signal. Our rationale is that 
a given writing system pertains to a specific form to encode a language, but cannot be considered to be part of the language itself.\footnote{As a matter of fact, languages existed well before writing was invented, and a given language can have different writing systems or new ones can be designed.}

\subsection{Multilinguality}

While most work in unsupervised cross-lingual learning considers two languages at a time, there have recently been some attempts to extend these methods to multiple languages \cite{Duong2017,Chen2018multilingual,Heyman2019}, and most work on unsupervised cross-lingual pre-training is multilingual \cite{Pires2019,conneau2019crosslingual}. When considering parallel data across a subset of the language pairs, multilinguality gives rise to additional scenarios. For instance, the scenario where two languages have no parallel data between each other but are well connected through a third (pivot) language has been explored by several authors in the context of machine translation \cite{cheng2016neural,Chen2017}. However, given that the languages in question are still indirectly connected through parallel data, this scenario does not fall within the \textit{unsupervised} category, and is instead commonly known as \textit{zero-resource} machine translation.

An alternative scenario explored in the contemporaneous work of \citet{liu2020multilingual} is where a set of languages are connected through parallel data, and there is a separate language with monolingual data only. We argue that, when it comes to the isolated language, such a scenario should still be considered as UCL, as it does not rely on any parallel data for that particular language nor does it assume any previous knowledge of it. This scenario is easy to justify from a practical perspective given the abundance of parallel data for high-resource languages, and can also be interesting from a more theoretical point of view. This way, rather than considering two unknown languages, this alternative scenario would assume some knowledge of how one particular language is connected to other languages, and attempt to align it to a separate unknown language.

\subsection{Discussion}

\begin{table}[t]
\resizebox{\columnwidth}{!}{%
\begin{tabular}{l l}
\toprule
Monolingual signal & Cross-lingual signal \\
\midrule
Sequence of symbols & Shared writing system \\
Sets of sentences/documents & Identical words \\
Tokens/subwords & String similarity \\
Linguistic analysis & \\
\bottomrule
\end{tabular}%
}
\caption{Different types of monolingual and cross-lingual signals that have been used for unsupervised cross-lingual learning, ordered roughly from least to most linguistic knowledge (top to bottom). }
\label{tab:monolingual-cross-lingual-signals}
\end{table}

As discussed throughout the section, there are different training signals that we can exploit depending on the available resources of the languages involved and the assumptions made regarding their writing system, which are summarized in Table~\ref{tab:monolingual-cross-lingual-signals}. Many of these signals are not specific to work on UCL but have been observed in the past in allegedly language-independent NLP approaches, as discussed by \citet{Bender2011}. Others, such as a reliance on subwords or shared symbols are more recent phenomena. 

While we do not aim to open a terminological debate on what UCL encompasses, we advocate for future work being more aware and explicit about the monolingual and cross-lingual signals they employ, what assumptions they make (e.g. regarding the writing system), and the extent to which these generalize to other languages.

In particular, we argue that it is critical to consider the assumptions made by different methods when comparing their results. Otherwise the blind chase for state-of-the-art performance may benefit models making stronger assumptions and exploiting all available training signals, which could ultimately conflict with the eminently scientific motivation of this research area (see \S\ref{subsec:motivation}).

\section{Methodological issues} \label{sec:methodological_issues}

In this section, we describe methodological issues that are commonly
encountered when training and evaluating unsupervised cross-lingual 
models and propose measures to ameliorate them.

\subsection{Validation and hyperparameter tuning}
In conventional supervised or semi-supervised settings, we use a separate validation set for development and hyperparameter tuning. However, this becomes tricky in unsupervised cross-lingual learning, where we ideally should not use any parallel data other than for testing purposes.

Previous work has not paid much attention to this aspect, and different methods are evaluated with different validation schemes. For instance, \citet{artetxe2018smt,artetxe2018nmt} use a \textbf{separate language pair} with a parallel validation set to make all development and hyperparameter decisions. They test their final system on other language pairs without any parallel data. This approach has the advantage of being strictly unsupervised with respect to the test language pairs, but the optimal hyperparameter choice might not necessarily transfer well across languages. 
In contrast, \citet{conneau2018word} and \citet{lample2018unsupervised} propose an \textbf{unsupervised validation criterion} that is defined over monolingual data and shown to correlate well with test performance. 
This enables systematic tuning on the language pair of interest, but still requires parallel data to guide the development of the unsupervised validation criterion itself. 
A \textbf{parallel validation set} has also been used for systematic tuning in the context of unsupervised machine translation \citep{marie2018unsupervised,marie2019nicts,stojanovski2019lmu}.
While this is motivated as a way to abstract away 
the issue of unsupervised tuning---which the authors consider to 
be an open problem---we argue that any systematic use of parallel 
data should not be considered UCL.
Finally, previous work often does not report the validation scheme used. 
In particular, unsupervised cross-lingual word embedding methods have almost exclusively been evaluated on bilingual lexicons 
that do not have a validation set, and presumably use the \textbf{test set} to guide development to some extent.

Our position is that a completely blind development model without any parallel data is unrealistic. Some cross-lingual signals to guide development are always needed. 
However, this factor should be carefully controlled and reported with the necessary rigor as a part of the experimental design. 
We advocate for using one language pair for development and evaluating on others
when possible.
If parallel data in the target language pair is used,
the test set should be kept blind to avoid overfitting,
and a separate validation should be used. %
In any case, we argue that the use of parallel data in the target language pair should be minimized if not completely avoided, and it should under no circumstances be used for extensive tuning. Instead, we recommend to use unsupervised validation criteria for systematic tuning in the target language.

\subsection{Evaluation practices}

We argue that there are also several issues with common evaluation practices in UCL.

\paragraph{Evaluation on favorable conditions.} Most work on UCL has focused on relatively close languages with large amounts of high-quality parallel corpora from similar domains.
Only recently have approaches considered more diverse languages as well as language pairs that do not involve English \cite{Glavas2019,Vulic2019}, and some existing methods have been shown to completely break in less favorable conditions \citep{guzman2019flores,marchisio2020unsupervised}. In addition, most approaches have focused on learning from similar domains, often involving Wikipedia and news corpora, which are unlikely to be available for low-resource languages.
We believe that future work should pay more attention to the effect of the typology and linguistic distance of the languages involved, as well as the size, noise and domain similarity of the training data used.%

\paragraph{Over-reliance on translation tasks.} Most work on UCL focuses on translation tasks, either at the word level (where the problem is known as \textit{bilingual lexicon induction}) or at the sentence level (where the problem is known as \textit{unsupervised machine translation}). While translation can be seen as the ultimate application of cross-lingual learning and has a strong practical interest on its own, it only evaluates a particular facet of a model's cross-lingual generalization ability. In relation to that, \citet{Glavas2019} showed that bilingual lexicon induction performance does not always correlate well with downstream tasks. In particular, they observe that some mapping methods that are specifically designed for bilingual lexicon induction perform poorly on other tasks, showing the risk of relying excessively on translation benchmarks for evaluating cross-lingual models.

Moreover, existing translation benchmarks have been shown to have several issues on their own.
In particular, bilingual lexicon induction datasets have been reported to misrepresent morphological variations, overly focus on named entities and frequent words, and have pervasive gaps in the gold-standard targets \cite{czarnowska2019dont,kementchedjhieva2019lost}. More generally, most of these datasets are limited to relatively close languages and comparable corpora.

\paragraph{Lack of an established cross-lingual benchmark.} At the same time, there is no \textit{de facto} standard benchmark to evaluate cross-lingual models beyond translation. Existing approaches have been evaluated in a wide variety of tasks including dependency parsing \cite{Schuster2019}, named entity recognition \cite{Rahimi2019}, sentiment analysis \citep{barnes2018bilingual}, natural language inference \cite{Conneau2018xnli}, and document classification \cite{Schwenk2018}. XNLI \cite{Conneau2018xnli} and MLDoc \cite{Schwenk2018} are common choices, but they have their own problems: MultiNLI, the dataset from which XNLI was derived, has been shown to contain superficial cues that can be exploited \cite{Gururangan2018}, while MLDoc can be solved by keyword matching \cite{Artetxe2019crosslingualtransfer}. There are non-English counterparts for more challenging tasks such as question answering \cite{Cui2019cross-lingual_rc,Hsu2019zero-shot_rc}, but these only exist for a handful of languages. More recent datasets such as XQuAD \cite{Artetxe2019crosslingualtransfer}, MLQA \cite{Lewis2019} and TyDi QA \citep{clark2020tydiqa} cover a wider set of languages, but a comprehensive benchmark that evaluates multilingual representations on a diverse set of tasks---in the style of GLUE \cite{Wang2019glue}---\emph{and} languages has been missing until very recently. The contemporaneous XTREME \cite{Hu2020xtreme} and XGLUE \citep{liang2020xglue} benchmarks try to close this gap, but they are still restricted to languages where existing labelled data is available.
Finally, an additional issue is that a large part of these benchmarks were created through translation, which was recently shown to introduce artifacts \citep{artetxe2020translation}.

\begin{table}[t]
\resizebox{\columnwidth}{!}{%
\begin{tabular}{ll}
\toprule
Methodological issues & Examples \\ \midrule \midrule
\begin{tabular}[c]{@{}l@{}}Validation and \\ hyperparameter tuning\end{tabular} & \begin{tabular}[c]{@{}l@{}}Systematic tuning with\\ parallel data or on test data\end{tabular} \\ \midrule
\begin{tabular}[c]{@{}l@{}}Evaluation on\\ favorable conditions\end{tabular} & \begin{tabular}[c]{@{}l@{}}Typologically similar languages; \\ always including English; \\ training on the same domain\end{tabular} \\ \midrule
\begin{tabular}[c]{@{}l@{}}Over-reliance on\\ translation tasks\end{tabular} & \begin{tabular}[c]{@{}l@{}}Overfitting to bilingual lexicon \\ induction; known issues with \\ existing datasets\end{tabular} \\ \midrule
\begin{tabular}[c]{@{}l@{}}Lack of an established\\ benchmark\end{tabular} & \begin{tabular}[c]{@{}l@{}}Evaluation on many different \\ tasks; problems with common\\ tasks (MLDoc and XNLI)\end{tabular} \\
\bottomrule
\end{tabular}%
}
\caption{Methodological issues pertaining to validation and hyperparameter tuning and evaluation practices in current work on unsupervised cross-lingual learning.}
\label{tab:methodological-issues}
\end{table}

We present a summary of the methodological issues discussed in Table \ref{tab:methodological-issues}.

\section{Bridging the gap between unsupervised cross-lingual learning flavors} \label{sec:unsupervised_types}

The three categories of UCL (\S\ref{sec:background}) have so far been treated as separate research topics by the community.
In particular, cross-lingual word embeddings have a long history \cite{Ruder2019survey}, while deep multilingual pretraining has emerged as a separate line of research with its own best practices and evaluation standards. At the same time, unsupervised machine translation has been considered a separate problem in its own right, where cross-lingual word embeddings and deep multilingual pretraining have just served as initialization techniques.

While each of these families have their own defining features, we believe that they share a strong connection that should be considered from a more holistic perspective. In particular, both cross-lingual word embeddings and deep multilingual pretraining share the goal of learning (sub)word representations, and essentially differ on whether such representations are static or context-dependent. Similarly, in addition to being a downstream application of the former, unsupervised machine translation can also be useful to develop other multilingual applications or learn better cross-lingual representations. This has previously been shown for \emph{supervised} machine translation \cite{Mccann2017,Siddhant2019evaluating} and recently for bilingual lexicon induction \citep{Artetxe2019bli}.
In light of these connections, we call for a more holistic view of UCL, both from an experimental and theoretical perspective.

\paragraph{Evaluation.} %
Most work on cross-lingual word embeddings focuses on bilingual lexicon induction. In contrast, deep multilingual pretraining has not been tested on this task, and is instead typically evaluated on zero-shot cross-lingual transfer. We think it is important to evaluate both families---cross-lingual word embeddings and deep multilingual representations---in the same conditions to better understand their strengths and weaknesses. In that regard, \citet{Artetxe2019crosslingualtransfer} recently showed that deep pretrained models are much stronger in some downstream tasks, while cross-lingual word embeddings are more efficient and sufficient for simpler tasks. However, this could partly be attributed to a particular integration strategy, and we advocate for using a common evaluation framework in future work to allow a direct comparison between the different families.

\paragraph{Theory.} From a more theoretical perspective, it is still not well understood in what ways cross-lingual word embeddings and deep multilingual pretraining differ. While one could expect the latter to be learning higher-level multilingual abstractions, recent work suggests that deep multilingual models might mostly be learning a lexical-level alignment \cite{Artetxe2019crosslingualtransfer}. For that reason, we believe that further research is needed to understand the relation between both families of models.

\section{Recommendations} 
\label{sec:recommendations}

To summarize, we make the following practical recommendations for future cross-lingual research: 
\begin{itemize}
\itemsep0em
    \item Be rigorous when motivating UCL. Do not present it as a practical scenario unless supported by a real use case.
    \item Be explicit about the monolingual and cross-lingual signals used by your approach and the assumptions it makes, and take them into considerations when comparing different models.
    \item Report the validation scheme used. Minimize the use of parallel data by preferring an unsupervised validation criterion and/or using only one language for development. Always keep the test set blind.
    \item Pay attention to the conditions in which you evaluate your model. Consider the impact of typology, linguistic distance, and the domain similarity, size and noise of the training data. Be aware of known issues with common benchmarks, and favor evaluation on a diverse set of tasks.
    \item Keep a holistic view of UCL, including cross-lingual word embeddings, deep multilingual pretraining and unsupervised machine translation. To the extent possible, favor a common evaluation framework for these different families.
\end{itemize}

\section{Conclusions}

In this position paper, we review the status quo of unsupervised cross-lingual learning---a relatively recent field. 
UCL is typically motivated by the lack of cross-lingual signal for many of the world's languages, but available resources indicate that a scenario with no parallel data and sufficient monolingual data is not realistic. Instead, we advocate for the importance of UCL for scientific reasons. 

We also discuss different monolingual and cross-lingual training signals that have been used in the past, and advocate for carefully reporting them to enable a meaningful comparison across different approaches.
In addition, we describe methodological issues related to the unsupervised setting
and propose measures to ameliorate them.  Finally, we discuss connections between cross-lingual word embeddings, deep multilingual pre-training, and unsupervised machine translation, calling for an evaluation on an equal footing.

We hope that this position paper will serve to strengthen research in UCL, providing a more rigorous look at the motivation, definition, and methodology.
In light of the unprecedented growth of our field in recent times, we believe that it is essential to establish a rigorous foundation connecting past and present research, and an evaluation protocol that carefully controls for the use of parallel data and assesses models in diverse, challenging settings.

\section*{Acknowledgments}

This research was partially funded by a Facebook Fellowship, the Basque Government excellence research group (IT1343-19), the Spanish MINECO (UnsupMT TIN2017‐91692‐EXP MCIU/AEI/FEDER, UE) and Project BigKnowledge (Ayudas Fundación BBVA a equipos de investigación científica 2018).

\bibliography{acl2020}

\begin{thebibliography}{87}
\expandafter\ifx\csname natexlab\endcsname\relax\def\natexlab#1{#1}\fi

\bibitem[{Agi{\'c} and Vuli{\'c}(2019)}]{agic2019jw300}
{\v{Z}}eljko Agi{\'c} and Ivan Vuli{\'c}. 2019.
\newblock \href {https://doi.org/10.18653/v1/P19-1310} {{JW}300: A
  wide-coverage parallel corpus for low-resource languages}.
\newblock In \emph{Proceedings of the 57th Annual Meeting of the Association
  for Computational Linguistics}, pages 3204--3210, Florence, Italy.
  Association for Computational Linguistics.

\bibitem[{Alvarez-Melis and Jaakkola(2018)}]{alvarezmelis2018gromov}
David Alvarez-Melis and Tommi Jaakkola. 2018.
\newblock \href {http://www.aclweb.org/anthology/D18-1214} {Gromov-wasserstein
  alignment of word embedding spaces}.
\newblock In \emph{Proceedings of the 2018 Conference on Empirical Methods in
  Natural Language Processing}, pages 1881--1890, Brussels, Belgium.
  Association for Computational Linguistics.

\bibitem[{Artetxe et~al.(2017)Artetxe, Labaka, and
  Agirre}]{artetxe2017learning}
Mikel Artetxe, Gorka Labaka, and Eneko Agirre. 2017.
\newblock \href {https://doi.org/10.18653/v1/P17-1042} {Learning bilingual word
  embeddings with (almost) no bilingual data}.
\newblock In \emph{Proceedings of the 55th Annual Meeting of the Association
  for Computational Linguistics (Volume 1: Long Papers)}, pages 451--462,
  Vancouver, Canada. Association for Computational Linguistics.

\bibitem[{Artetxe et~al.(2018{\natexlab{a}})Artetxe, Labaka, and
  Agirre}]{artetxe2018robust}
Mikel Artetxe, Gorka Labaka, and Eneko Agirre. 2018{\natexlab{a}}.
\newblock \href {https://doi.org/10.18653/v1/P18-1073} {A robust self-learning
  method for fully unsupervised cross-lingual mappings of word embeddings}.
\newblock In \emph{Proceedings of the 56th Annual Meeting of the Association
  for Computational Linguistics (Volume 1: Long Papers)}, pages 789--798,
  Melbourne, Australia. Association for Computational Linguistics.

\bibitem[{Artetxe et~al.(2018{\natexlab{b}})Artetxe, Labaka, and
  Agirre}]{artetxe2018smt}
Mikel Artetxe, Gorka Labaka, and Eneko Agirre. 2018{\natexlab{b}}.
\newblock \href {https://doi.org/10.18653/v1/D18-1399} {Unsupervised
  statistical machine translation}.
\newblock In \emph{Proceedings of the 2018 Conference on Empirical Methods in
  Natural Language Processing}, pages 3632--3642, Brussels, Belgium.
  Association for Computational Linguistics.

\bibitem[{Artetxe et~al.(2019{\natexlab{a}})Artetxe, Labaka, and
  Agirre}]{Artetxe2019bli}
Mikel Artetxe, Gorka Labaka, and Eneko Agirre. 2019{\natexlab{a}}.
\newblock \href {https://doi.org/10.18653/v1/P19-1494} {Bilingual lexicon
  induction through unsupervised machine translation}.
\newblock In \emph{Proceedings of the 57th Annual Meeting of the Association
  for Computational Linguistics}, pages 5002--5007, Florence, Italy.
  Association for Computational Linguistics.

\bibitem[{Artetxe et~al.(2019{\natexlab{b}})Artetxe, Labaka, and
  Agirre}]{artetxe2019effective}
Mikel Artetxe, Gorka Labaka, and Eneko Agirre. 2019{\natexlab{b}}.
\newblock \href {https://doi.org/10.18653/v1/P19-1019} {An effective approach
  to unsupervised machine translation}.
\newblock In \emph{Proceedings of the 57th Annual Meeting of the Association
  for Computational Linguistics}, pages 194--203, Florence, Italy. Association
  for Computational Linguistics.

\bibitem[{Artetxe et~al.(2020{\natexlab{a}})Artetxe, Labaka, and
  Agirre}]{artetxe2020translation}
Mikel Artetxe, Gorka Labaka, and Eneko Agirre. 2020{\natexlab{a}}.
\newblock \href {https://arxiv.org/abs/2004.04721} {Translation artifacts in
  cross-lingual transfer learning}.
\newblock \emph{arXiv preprint arXiv:2004.04721}.

\bibitem[{Artetxe et~al.(2018{\natexlab{c}})Artetxe, Labaka, Agirre, and
  Cho}]{artetxe2018nmt}
Mikel Artetxe, Gorka Labaka, Eneko Agirre, and Kyunghyun Cho.
  2018{\natexlab{c}}.
\newblock \href {https://openreview.net/pdf?id=Sy2ogebAW} {Unsupervised neural
  machine translation}.
\newblock In \emph{Proceedings of the 6th International Conference on Learning
  Representations (ICLR 2018)}.

\bibitem[{Artetxe et~al.(2020{\natexlab{b}})Artetxe, Ruder, and
  Yogatama}]{Artetxe2019crosslingualtransfer}
Mikel Artetxe, Sebastian Ruder, and Dani Yogatama. 2020{\natexlab{b}}.
\newblock {On the Cross-lingual Transferability of Monolingual
  Representations}.
\newblock In \emph{Proceedings of ACL 2020}.

\bibitem[{Barnes et~al.(2018)Barnes, Klinger, and Schulte~im
  Walde}]{barnes2018bilingual}
Jeremy Barnes, Roman Klinger, and Sabine Schulte~im Walde. 2018.
\newblock \href {https://doi.org/10.18653/v1/P18-1231} {Bilingual sentiment
  embeddings: Joint projection of sentiment across languages}.
\newblock In \emph{Proceedings of the 56th Annual Meeting of the Association
  for Computational Linguistics (Volume 1: Long Papers)}, pages 2483--2493,
  Melbourne, Australia. Association for Computational Linguistics.

\bibitem[{Bender(2011)}]{Bender2011}
Emily~M. Bender. 2011.
\newblock {On Achieving and Evaluating Language-Independence in NLP}.
\newblock \emph{Linguistic Issues in Language Technology}, 6(3):1--26.

\bibitem[{Bojanowski et~al.(2017)Bojanowski, Grave, Joulin, and
  Mikolov}]{Bojanowski2017}
Piotr Bojanowski, Edouard Grave, Armand Joulin, and Tomas Mikolov. 2017.
\newblock \href {https://doi.org/10.1162/tacl_a_00051} {Enriching word vectors
  with subword information}.
\newblock \emph{Transactions of the Association for Computational Linguistics},
  5:135--146.

\bibitem[{Buck et~al.(2014)Buck, Heafield, and van Ooyen}]{buck2014ngrams}
Christian Buck, Kenneth Heafield, and Bas van Ooyen. 2014.
\newblock \href
  {http://www.lrec-conf.org/proceedings/lrec2014/pdf/1097_Paper.pdf} {N-gram
  counts and language models from the common crawl}.
\newblock In \emph{Proceedings of the Ninth International Conference on
  Language Resources and Evaluation ({LREC}'14)}, pages 3579--3584, Reykjavik,
  Iceland. European Language Resources Association (ELRA).

\bibitem[{Chen and Cardie(2018)}]{Chen2018multilingual}
Xilun Chen and Claire Cardie. 2018.
\newblock \href {https://doi.org/10.18653/v1/D18-1024} {Unsupervised
  multilingual word embeddings}.
\newblock In \emph{Proceedings of the 2018 Conference on Empirical Methods in
  Natural Language Processing}, pages 261--270, Brussels, Belgium. Association
  for Computational Linguistics.

\bibitem[{Chen et~al.(2017)Chen, Liu, Cheng, and Li}]{Chen2017}
Yun Chen, Yang Liu, Yong Cheng, and Victor~O.K. Li. 2017.
\newblock \href {https://doi.org/10.18653/v1/P17-1176} {A teacher-student
  framework for zero-resource neural machine translation}.
\newblock In \emph{Proceedings of the 55th Annual Meeting of the Association
  for Computational Linguistics (Volume 1: Long Papers)}, pages 1925--1935,
  Vancouver, Canada. Association for Computational Linguistics.

\bibitem[{Cheng et~al.(2016)Cheng, Liu, Yang, Sun, and Xu}]{cheng2016neural}
Yong Cheng, Yang Liu, Qian Yang, Maosong Sun, and Wei Xu. 2016.
\newblock \href {https://arxiv.org/abs/1611.04928} {Neural machine translation
  with pivot languages}.
\newblock \emph{arXiv preprint arXiv:1611.04928}.

\bibitem[{Clark et~al.(2020)Clark, Choi, Collins, Garrette, Kwiatkowski,
  Nikolaev, and Palomaki}]{clark2020tydiqa}
Jonathan~H. Clark, Eunsol Choi, Michael Collins, Dan Garrette, Tom Kwiatkowski,
  Vitaly Nikolaev, and Jennimaria Palomaki. 2020.
\newblock Tydi qa: A benchmark for information-seeking question answering in
  typologically diverse languages.
\newblock \emph{Transactions of the Association for Computational Linguistics}.

\bibitem[{Conneau et~al.(2019)Conneau, Khandelwal, Goyal, Chaudhary, Wenzek,
  Guzm{\'a}n, Grave, Ott, Zettlemoyer, and Stoyanov}]{conneau2019unsupervised}
Alexis Conneau, Kartikay Khandelwal, Naman Goyal, Vishrav Chaudhary, Guillaume
  Wenzek, Francisco Guzm{\'a}n, Edouard Grave, Myle Ott, Luke Zettlemoyer, and
  Veselin Stoyanov. 2019.
\newblock \href {https://arxiv.org/abs/1911.02116} {Unsupervised cross-lingual
  representation learning at scale}.
\newblock \emph{arXiv preprint arXiv:1911.02116}.

\bibitem[{Conneau and Lample(2019)}]{conneau2019crosslingual}
Alexis Conneau and Guillaume Lample. 2019.
\newblock \href
  {http://papers.nips.cc/paper/8928-cross-lingual-language-model-pretraining.pdf}
  {Cross-lingual language model pretraining}.
\newblock In \emph{Advances in Neural Information Processing Systems 32}, pages
  7057--7067.

\bibitem[{Conneau et~al.(2018{\natexlab{a}})Conneau, Lample, Ranzato, Denoyer,
  and J{\'{e}}gou}]{conneau2018word}
Alexis Conneau, Guillaume Lample, Marc'Aurelio Ranzato, Ludovic Denoyer, and
  Herv{\'{e}} J{\'{e}}gou. 2018{\natexlab{a}}.
\newblock \href {https://openreview.net/pdf?id=H196sainb} {Word translation
  without parallel data}.
\newblock In \emph{Proceedings of the 6th International Conference on Learning
  Representations (ICLR 2018)}.

\bibitem[{Conneau et~al.(2018{\natexlab{b}})Conneau, Rinott, Lample, Williams,
  Bowman, Schwenk, and Stoyanov}]{Conneau2018xnli}
Alexis Conneau, Ruty Rinott, Guillaume Lample, Adina Williams, Samuel Bowman,
  Holger Schwenk, and Veselin Stoyanov. 2018{\natexlab{b}}.
\newblock \href {https://doi.org/10.18653/v1/D18-1269} {{XNLI}: Evaluating
  cross-lingual sentence representations}.
\newblock In \emph{Proceedings of the 2018 Conference on Empirical Methods in
  Natural Language Processing}, pages 2475--2485, Brussels, Belgium.
  Association for Computational Linguistics.

\bibitem[{Cui et~al.(2019)Cui, Che, Liu, Qin, Wang, and
  Hu}]{Cui2019cross-lingual_rc}
Yiming Cui, Wanxiang Che, Ting Liu, Bing Qin, Shijin Wang, and Guoping Hu.
  2019.
\newblock \href {https://doi.org/10.18653/v1/D19-1169} {Cross-lingual machine
  reading comprehension}.
\newblock In \emph{Proceedings of the 2019 Conference on Empirical Methods in
  Natural Language Processing and the 9th International Joint Conference on
  Natural Language Processing (EMNLP-IJCNLP)}, pages 1586--1595, Hong Kong,
  China. Association for Computational Linguistics.

\bibitem[{Czarnowska et~al.(2019)Czarnowska, Ruder, Grave, Cotterell, and
  Copestake}]{czarnowska2019dont}
Paula Czarnowska, Sebastian Ruder, Edouard Grave, Ryan Cotterell, and Ann
  Copestake. 2019.
\newblock \href {https://doi.org/10.18653/v1/D19-1090} {Don{'}t forget the long
  tail! {A} comprehensive analysis of morphological generalization in bilingual
  lexicon induction}.
\newblock In \emph{Proceedings of the 2019 Conference on Empirical Methods in
  Natural Language Processing and the 9th International Joint Conference on
  Natural Language Processing (EMNLP-IJCNLP)}, pages 973--982, Hong Kong,
  China. Association for Computational Linguistics.

\bibitem[{Dai and Le(2015)}]{dai2015semisupervised}
Andrew~M. Dai and Quoc~V. Le. 2015.
\newblock \href
  {http://papers.nips.cc/paper/5949-semi-supervised-sequence-learning.pdf}
  {Semi-supervised sequence learning}.
\newblock In \emph{Advances in Neural Information Processing Systems 28}, pages
  3079--3087.

\bibitem[{Devlin et~al.(2019)Devlin, Chang, Lee, and
  Toutanova}]{devlin2018bert}
Jacob Devlin, Ming-Wei Chang, Kenton Lee, and Kristina Toutanova. 2019.
\newblock \href {https://www.aclweb.org/anthology/N19-1423} {{BERT}:
  Pre-training of deep bidirectional transformers for language understanding}.
\newblock In \emph{Proceedings of the 2019 Conference of the North {A}merican
  Chapter of the Association for Computational Linguistics: Human Language
  Technologies, Volume 1 (Long and Short Papers)}, pages 4171--4186,
  Minneapolis, Minnesota. Association for Computational Linguistics.

\bibitem[{Dou and Knight(2012)}]{dou2012large}
Qing Dou and Kevin Knight. 2012.
\newblock \href {https://www.aclweb.org/anthology/D12-1025} {Large scale
  decipherment for out-of-domain machine translation}.
\newblock In \emph{Proceedings of the 2012 Joint Conference on Empirical
  Methods in Natural Language Processing and Computational Natural Language
  Learning}, pages 266--275, Jeju Island, Korea. Association for Computational
  Linguistics.

\bibitem[{Dou and Knight(2013)}]{dou2013dependency}
Qing Dou and Kevin Knight. 2013.
\newblock \href {https://www.aclweb.org/anthology/D13-1173} {Dependency-based
  decipherment for resource-limited machine translation}.
\newblock In \emph{Proceedings of the 2013 Conference on Empirical Methods in
  Natural Language Processing}, pages 1668--1676, Seattle, Washington, USA.
  Association for Computational Linguistics.

\bibitem[{Duong et~al.(2017)Duong, Kanayama, Ma, Bird, and Cohn}]{Duong2017}
Long Duong, Hiroshi Kanayama, Tengfei Ma, Steven Bird, and Trevor Cohn. 2017.
\newblock \href {https://www.aclweb.org/anthology/E17-1084} {Multilingual
  training of crosslingual word embeddings}.
\newblock In \emph{Proceedings of the 15th Conference of the {E}uropean Chapter
  of the Association for Computational Linguistics: Volume 1, Long Papers},
  pages 894--904, Valencia, Spain. Association for Computational Linguistics.

\bibitem[{Eco and Fentress(1995)}]{eco1995search}
Umberto Eco and James Fentress. 1995.
\newblock \emph{The search for the perfect language}.
\newblock Blackwell Oxford.

\bibitem[{Faruqui and Dyer(2014)}]{faruqui2014improving}
Manaal Faruqui and Chris Dyer. 2014.
\newblock \href {https://doi.org/10.3115/v1/E14-1049} {Improving vector space
  word representations using multilingual correlation}.
\newblock In \emph{Proceedings of the 14th Conference of the {E}uropean Chapter
  of the Association for Computational Linguistics}, pages 462--471,
  Gothenburg, Sweden. Association for Computational Linguistics.

\bibitem[{Glava{\v{s}} et~al.(2019)Glava{\v{s}}, Litschko, Ruder, and
  Vuli{\'c}}]{Glavas2019}
Goran Glava{\v{s}}, Robert Litschko, Sebastian Ruder, and Ivan Vuli{\'c}. 2019.
\newblock \href {https://doi.org/10.18653/v1/P19-1070} {How to (properly)
  evaluate cross-lingual word embeddings: On strong baselines, comparative
  analyses, and some misconceptions}.
\newblock In \emph{Proceedings of the 57th Annual Meeting of the Association
  for Computational Linguistics}, pages 710--721, Florence, Italy. Association
  for Computational Linguistics.

\bibitem[{Gouws et~al.(2015)Gouws, Bengio, and Corrado}]{gouws2015bilbowa}
Stephan Gouws, Yoshua Bengio, and Greg Corrado. 2015.
\newblock \href {http://proceedings.mlr.press/v37/gouws15.html} {{BilBOWA}:
  Fast bilingual distributed representations without word alignments}.
\newblock In \emph{Proceedings of the 32nd International Conference on Machine
  Learning}, volume~37 of \emph{Proceedings of Machine Learning Research},
  pages 748--756, Lille, France. PMLR.

\bibitem[{Grave et~al.(2018)Grave, Bojanowski, Gupta, Joulin, and
  Mikolov}]{Grave2018}
Edouard Grave, Piotr Bojanowski, Prakhar Gupta, Armand Joulin, and Tomas
  Mikolov. 2018.
\newblock \href {https://www.aclweb.org/anthology/L18-1550} {Learning word
  vectors for 157 languages}.
\newblock In \emph{Proceedings of the Eleventh International Conference on
  Language Resources and Evaluation ({LREC} 2018)}, Miyazaki, Japan. European
  Language Resources Association (ELRA).

\bibitem[{Grave et~al.(2019)Grave, Joulin, and Berthet}]{grave2018unsupervised}
Edouard Grave, Armand Joulin, and Quentin Berthet. 2019.
\newblock \href {http://proceedings.mlr.press/v89/grave19a.html} {Unsupervised
  alignment of embeddings with wasserstein procrustes}.
\newblock In \emph{Proceedings of Machine Learning Research}, volume~89, pages
  1880--1890. PMLR.

\bibitem[{Gururangan et~al.(2018)Gururangan, Swayamdipta, Levy, Schwartz,
  Bowman, and Smith}]{Gururangan2018}
Suchin Gururangan, Swabha Swayamdipta, Omer Levy, Roy Schwartz, Samuel Bowman,
  and Noah~A. Smith. 2018.
\newblock \href {https://doi.org/10.18653/v1/N18-2017} {Annotation artifacts in
  natural language inference data}.
\newblock In \emph{Proceedings of the 2018 Conference of the North {A}merican
  Chapter of the Association for Computational Linguistics: Human Language
  Technologies, Volume 2 (Short Papers)}, pages 107--112, New Orleans,
  Louisiana. Association for Computational Linguistics.

\bibitem[{Guzm{\'a}n et~al.(2019)Guzm{\'a}n, Chen, Ott, Pino, Lample, Koehn,
  Chaudhary, and Ranzato}]{guzman2019flores}
Francisco Guzm{\'a}n, Peng-Jen Chen, Myle Ott, Juan Pino, Guillaume Lample,
  Philipp Koehn, Vishrav Chaudhary, and Marc{'}Aurelio Ranzato. 2019.
\newblock \href {https://doi.org/10.18653/v1/D19-1632} {The {FLORES} evaluation
  datasets for low-resource machine translation: {N}epali{--}{E}nglish and
  {S}inhala{--}{E}nglish}.
\newblock In \emph{Proceedings of the 2019 Conference on Empirical Methods in
  Natural Language Processing and the 9th International Joint Conference on
  Natural Language Processing (EMNLP-IJCNLP)}, pages 6097--6110, Hong Kong,
  China. Association for Computational Linguistics.

\bibitem[{Hahn and Baroni(2019)}]{Hahn2019tabularasa}
Michael Hahn and Marco Baroni. 2019.
\newblock \href {https://doi.org/10.1162/tacl\_a\_00283} {Tabula nearly rasa:
  Probing the linguistic knowledge of character-level neural language models
  trained on unsegmented text}.
\newblock \emph{Transactions of the Association for Computational Linguistics},
  7:467--484.

\bibitem[{Heyman et~al.(2019)Heyman, Verreet, Vuli{\'c}, and
  Moens}]{Heyman2019}
Geert Heyman, Bregt Verreet, Ivan Vuli{\'c}, and Marie-Francine Moens. 2019.
\newblock \href {https://doi.org/10.18653/v1/N19-1188} {Learning unsupervised
  multilingual word embeddings with incremental multilingual hubs}.
\newblock In \emph{Proceedings of the 2019 Conference of the North {A}merican
  Chapter of the Association for Computational Linguistics: Human Language
  Technologies, Volume 1 (Long and Short Papers)}, pages 1890--1902,
  Minneapolis, Minnesota. Association for Computational Linguistics.

\bibitem[{Heyman et~al.(2017)Heyman, Vuli{\'c}, and Moens}]{Heyman2017}
Geert Heyman, Ivan Vuli{\'c}, and Marie-Francine Moens. 2017.
\newblock \href {https://www.aclweb.org/anthology/E17-1102} {Bilingual lexicon
  induction by learning to combine word-level and character-level
  representations}.
\newblock In \emph{Proceedings of the 15th Conference of the {E}uropean Chapter
  of the Association for Computational Linguistics: Volume 1, Long Papers},
  pages 1085--1095, Valencia, Spain. Association for Computational Linguistics.

\bibitem[{Hoshen and Wolf(2018)}]{hoshen2018nonadversarial}
Yedid Hoshen and Lior Wolf. 2018.
\newblock \href {https://doi.org/10.18653/v1/D18-1043} {Non-adversarial
  unsupervised word translation}.
\newblock In \emph{Proceedings of the 2018 Conference on Empirical Methods in
  Natural Language Processing}, pages 469--478, Brussels, Belgium. Association
  for Computational Linguistics.

\bibitem[{Howard and Ruder(2018)}]{howard2018universal}
Jeremy Howard and Sebastian Ruder. 2018.
\newblock \href {https://www.aclweb.org/anthology/P18-1031} {Universal language
  model fine-tuning for text classification}.
\newblock In \emph{Proceedings of the 56th Annual Meeting of the Association
  for Computational Linguistics (Volume 1: Long Papers)}, pages 328--339,
  Melbourne, Australia. Association for Computational Linguistics.

\bibitem[{Hsu et~al.(2019)Hsu, Liu, and Lee}]{Hsu2019zero-shot_rc}
Tsung-Yuan Hsu, Chi-Liang Liu, and Hung-yi Lee. 2019.
\newblock \href {https://doi.org/10.18653/v1/D19-1607} {Zero-shot reading
  comprehension by cross-lingual transfer learning with multi-lingual language
  representation model}.
\newblock In \emph{Proceedings of the 2019 Conference on Empirical Methods in
  Natural Language Processing and the 9th International Joint Conference on
  Natural Language Processing (EMNLP-IJCNLP)}, pages 5933--5940, Hong Kong,
  China. Association for Computational Linguistics.

\bibitem[{Hu et~al.(2020)Hu, Ruder, Siddhant, Neubig, Firat, and
  Johnson}]{Hu2020xtreme}
Junjie Hu, Sebastian Ruder, Aditya Siddhant, Graham Neubig, Orhan Firat, and
  Melvin Johnson. 2020.
\newblock \href {https://arxiv.org/abs/2003.11080} {{XTREME: A Massively
  Multilingual Multi-task Benchmark for Evaluating Cross-lingual
  Generalization}}.
\newblock \emph{arXiv preprint arXiv:2003.11080}.

\bibitem[{Kamholz et~al.(2014)Kamholz, Pool, and Colowick}]{kamholz2014panlex}
David Kamholz, Jonathan Pool, and Susan Colowick. 2014.
\newblock \href
  {http://www.lrec-conf.org/proceedings/lrec2014/pdf/1029_Paper.pdf}
  {{P}an{L}ex: Building a resource for panlingual lexical translation}.
\newblock In \emph{Proceedings of the Ninth International Conference on
  Language Resources and Evaluation ({LREC}'14)}, pages 3145--3150, Reykjavik,
  Iceland. European Language Resources Association (ELRA).

\bibitem[{Kementchedjhieva et~al.(2019)Kementchedjhieva, Hartmann, and
  S{\o}gaard}]{kementchedjhieva2019lost}
Yova Kementchedjhieva, Mareike Hartmann, and Anders S{\o}gaard. 2019.
\newblock \href {https://doi.org/10.18653/v1/D19-1328} {Lost in evaluation:
  Misleading benchmarks for bilingual dictionary induction}.
\newblock In \emph{Proceedings of the 2019 Conference on Empirical Methods in
  Natural Language Processing and the 9th International Joint Conference on
  Natural Language Processing (EMNLP-IJCNLP)}, pages 3327--3332, Hong Kong,
  China. Association for Computational Linguistics.

\bibitem[{Lample et~al.(2018{\natexlab{a}})Lample, Conneau, Denoyer, and
  Ranzato}]{lample2018unsupervised}
Guillaume Lample, Alexis Conneau, Ludovic Denoyer, and Marc'Aurelio Ranzato.
  2018{\natexlab{a}}.
\newblock \href {https://openreview.net/pdf?id=rkYTTf-AZ} {Unsupervised machine
  translation using monolingual corpora only}.
\newblock In \emph{Proceedings of the 6th International Conference on Learning
  Representations (ICLR 2018)}.

\bibitem[{Lample et~al.(2018{\natexlab{b}})Lample, Ott, Conneau, Denoyer, and
  Ranzato}]{lample2018phrase}
Guillaume Lample, Myle Ott, Alexis Conneau, Ludovic Denoyer, and Marc{'}Aurelio
  Ranzato. 2018{\natexlab{b}}.
\newblock \href {https://doi.org/10.18653/v1/D18-1549} {Phrase-based {\&}
  neural unsupervised machine translation}.
\newblock In \emph{Proceedings of the 2018 Conference on Empirical Methods in
  Natural Language Processing}, pages 5039--5049, Brussels, Belgium.
  Association for Computational Linguistics.

\bibitem[{Lewis et~al.(2019)Lewis, Oğuz, Rinott, Riedel, and
  Schwenk}]{Lewis2019}
Patrick Lewis, Barlas Oğuz, Ruty Rinott, Sebastian Riedel, and Holger Schwenk.
  2019.
\newblock \href {https://arxiv.org/abs/1910.07475} {{MLQA: Evaluating
  Cross-lingual Extractive Question Answering}}.
\newblock \emph{arXiv preprint arXiv:1910.07475}.

\bibitem[{Liang et~al.(2020)Liang, Duan, Gong, Wu, Guo, Qi, Gong, Shou, Jiang,
  Cao, Fan, Zhang, Agrawal, Cui, Wei, Bharti, Qiao, Chen, Wu, Liu, Yang,
  Majumder, and Zhou}]{liang2020xglue}
Yaobo Liang, Nan Duan, Yeyun Gong, Ning Wu, Fenfei Guo, Weizhen Qi, Ming Gong,
  Linjun Shou, Daxin Jiang, Guihong Cao, Xiaodong Fan, Bruce Zhang, Rahul
  Agrawal, Edward Cui, Sining Wei, Taroon Bharti, Ying Qiao, Jiun-Hung Chen,
  Winnie Wu, Shuguang Liu, Fan Yang, Rangan Majumder, and Ming Zhou. 2020.
\newblock \href {https://arxiv.org/abs/2004.01401} {Xglue: A new benchmark
  dataset for cross-lingual pre-training, understanding and generation}.
\newblock \emph{arXiv preprint arXiv:2004.01401}.

\bibitem[{Liu et~al.(2020)Liu, Gu, Goyal, Li, Edunov, Ghazvininejad, Lewis, and
  Zettlemoyer}]{liu2020multilingual}
Yinhan Liu, Jiatao Gu, Naman Goyal, Xian Li, Sergey Edunov, Marjan
  Ghazvininejad, Mike Lewis, and Luke Zettlemoyer. 2020.
\newblock \href {https://arxiv.org/abs/2001.08210} {Multilingual denoising
  pre-training for neural machine translation}.
\newblock \emph{arXiv preprint arXiv:2001.08210}.

\bibitem[{Luong et~al.(2015)Luong, Pham, and Manning}]{luong2015bilingual}
Thang Luong, Hieu Pham, and Christopher~D. Manning. 2015.
\newblock \href {https://doi.org/10.3115/v1/W15-1521} {Bilingual word
  representations with monolingual quality in mind}.
\newblock In \emph{Proceedings of the 1st Workshop on Vector Space Modeling for
  Natural Language Processing}, pages 151--159, Denver, Colorado. Association
  for Computational Linguistics.

\bibitem[{Marchisio et~al.(2020)Marchisio, Duh, and
  Koehn}]{marchisio2020unsupervised}
Kelly Marchisio, Kevin Duh, and Philipp Koehn. 2020.
\newblock \href {https://arxiv.org/abs/2004.05516} {When does unsupervised
  machine translation work?}
\newblock \emph{arXiv preprint arXiv:2004.05516}.

\bibitem[{Marie and Fujita(2018)}]{marie2018unsupervised}
Benjamin Marie and Atsushi Fujita. 2018.
\newblock \href {https://arxiv.org/abs/1810.12703} {Unsupervised neural machine
  translation initialized by unsupervised statistical machine translation}.
\newblock \emph{arXiv preprint arXiv:1810.12703}.

\bibitem[{Marie et~al.(2019)Marie, Sun, Wang, Chen, Fujita, Utiyama, and
  Sumita}]{marie2019nicts}
Benjamin Marie, Haipeng Sun, Rui Wang, Kehai Chen, Atsushi Fujita, Masao
  Utiyama, and Eiichiro Sumita. 2019.
\newblock \href {https://doi.org/10.18653/v1/W19-5330} {{NICT}{'}s unsupervised
  neural and statistical machine translation systems for the {WMT}19 news
  translation task}.
\newblock In \emph{Proceedings of the Fourth Conference on Machine Translation
  (Volume 2: Shared Task Papers, Day 1)}, pages 294--301, Florence, Italy.
  Association for Computational Linguistics.

\bibitem[{Mayer and Cysouw(2014)}]{mayer2014creating}
Thomas Mayer and Michael Cysouw. 2014.
\newblock \href
  {http://www.lrec-conf.org/proceedings/lrec2014/pdf/220_Paper.pdf} {Creating a
  massively parallel {B}ible corpus}.
\newblock In \emph{Proceedings of the Ninth International Conference on
  Language Resources and Evaluation ({LREC}'14)}, pages 3158--3163, Reykjavik,
  Iceland. European Language Resources Association (ELRA).

\bibitem[{McCann et~al.(2017)McCann, Bradbury, Xiong, and Socher}]{Mccann2017}
Bryan McCann, James Bradbury, Caiming Xiong, and Richard Socher. 2017.
\newblock \href
  {http://papers.nips.cc/paper/7209-learned-in-translation-contextualized-word-vectors.pdf}
  {Learned in translation: Contextualized word vectors}.
\newblock In \emph{Advances in Neural Information Processing Systems 30}, pages
  6294--6305.

\bibitem[{Mikolov et~al.(2013{\natexlab{a}})Mikolov, Le, and
  Sutskever}]{mikolov2013exploiting}
Tomas Mikolov, Quoc~V Le, and Ilya Sutskever. 2013{\natexlab{a}}.
\newblock \href {https://arxiv.org/abs/1309.4168} {Exploiting similarities
  among languages for machine translation}.
\newblock \emph{arXiv preprint arXiv:1309.4168}.

\bibitem[{Mikolov et~al.(2013{\natexlab{b}})Mikolov, Sutskever, Chen, Corrado,
  and Dean}]{mikolov2013distributed}
Tomas Mikolov, Ilya Sutskever, Kai Chen, Greg~S. Corrado, and Jeff Dean.
  2013{\natexlab{b}}.
\newblock \href
  {http://papers.nips.cc/paper/5021-distributed-representations-of-words-and-phrases-and-their-compositionality.pdf}
  {Distributed representations of words and phrases and their
  compositionality}.
\newblock In \emph{Advances in Neural Information Processing Systems 26}, pages
  3111--3119.

\bibitem[{Pennington et~al.(2014)Pennington, Socher, and
  Manning}]{pennington2014glove}
Jeffrey Pennington, Richard Socher, and Christopher Manning. 2014.
\newblock \href {https://doi.org/10.3115/v1/D14-1162} {{G}lo{V}e: Global
  vectors for word representation}.
\newblock In \emph{Proceedings of the 2014 Conference on Empirical Methods in
  Natural Language Processing ({EMNLP})}, pages 1532--1543, Doha, Qatar.
  Association for Computational Linguistics.

\bibitem[{Peters et~al.(2018)Peters, Neumann, Iyyer, Gardner, Clark, Lee, and
  Zettlemoyer}]{peters2018deep}
Matthew Peters, Mark Neumann, Mohit Iyyer, Matt Gardner, Christopher Clark,
  Kenton Lee, and Luke Zettlemoyer. 2018.
\newblock \href {https://doi.org/10.18653/v1/N18-1202} {Deep contextualized
  word representations}.
\newblock In \emph{Proceedings of the 2018 Conference of the North {A}merican
  Chapter of the Association for Computational Linguistics: Human Language
  Technologies, Volume 1 (Long Papers)}, pages 2227--2237, New Orleans,
  Louisiana. Association for Computational Linguistics.

\bibitem[{Pires et~al.(2019)Pires, Schlinger, and Garrette}]{Pires2019}
Telmo Pires, Eva Schlinger, and Dan Garrette. 2019.
\newblock \href {https://doi.org/10.18653/v1/P19-1493} {How multilingual is
  multilingual {BERT}?}
\newblock In \emph{Proceedings of the 57th Annual Meeting of the Association
  for Computational Linguistics}, pages 4996--5001, Florence, Italy.
  Association for Computational Linguistics.

\bibitem[{Rahimi et~al.(2019)Rahimi, Li, and Cohn}]{Rahimi2019}
Afshin Rahimi, Yuan Li, and Trevor Cohn. 2019.
\newblock \href {https://doi.org/10.18653/v1/P19-1015} {Massively multilingual
  transfer for {NER}}.
\newblock In \emph{Proceedings of the 57th Annual Meeting of the Association
  for Computational Linguistics}, pages 151--164, Florence, Italy. Association
  for Computational Linguistics.

\bibitem[{Ravi and Knight(2011)}]{ravi2011deciphering}
Sujith Ravi and Kevin Knight. 2011.
\newblock \href {https://www.aclweb.org/anthology/P11-1002} {Deciphering
  foreign language}.
\newblock In \emph{Proceedings of the 49th Annual Meeting of the Association
  for Computational Linguistics: Human Language Technologies}, pages 12--21,
  Portland, Oregon, USA. Association for Computational Linguistics.

\bibitem[{Ren et~al.(2019)Ren, Zhang, Liu, Zhou, and Ma}]{ren2019unsupervised}
Shuo Ren, Zhirui Zhang, Shujie Liu, Ming Zhou, and Shuai Ma. 2019.
\newblock Unsupervised neural machine translation with {SMT} as posterior
  regularization.
\newblock In \emph{Proceedings of the AAAI Conference on Artificial
  Intelligence}, volume~33, pages 241--248.

\bibitem[{Riley and Gildea(2018)}]{Riley2018}
Parker Riley and Daniel Gildea. 2018.
\newblock \href {https://doi.org/10.18653/v1/P18-2062} {Orthographic features
  for bilingual lexicon induction}.
\newblock In \emph{Proceedings of the 56th Annual Meeting of the Association
  for Computational Linguistics (Volume 2: Short Papers)}, pages 390--394,
  Melbourne, Australia. Association for Computational Linguistics.

\bibitem[{Ruder et~al.(2018)Ruder, Cotterell, Kementchedjhieva, and
  S{\o}gaard}]{Ruder2018discriminative}
Sebastian Ruder, Ryan Cotterell, Yova Kementchedjhieva, and Anders S{\o}gaard.
  2018.
\newblock \href {https://doi.org/10.18653/v1/D18-1042} {A discriminative
  latent-variable model for bilingual lexicon induction}.
\newblock In \emph{Proceedings of the 2018 Conference on Empirical Methods in
  Natural Language Processing}, pages 458--468, Brussels, Belgium. Association
  for Computational Linguistics.

\bibitem[{Ruder et~al.(2019)Ruder, Vuli{\'{c}}, and
  S{\o}gaard}]{Ruder2019survey}
Sebastian Ruder, Ivan Vuli{\'{c}}, and Anders S{\o}gaard. 2019.
\newblock {A Survey of Cross-lingual Word Embedding Models}.
\newblock \emph{Journal of Artificial Intelligence Research}, 65:569--631.

\bibitem[{Schuster et~al.(2019)Schuster, Ram, Barzilay, and
  Globerson}]{Schuster2019}
Tal Schuster, Ori Ram, Regina Barzilay, and Amir Globerson. 2019.
\newblock \href {https://doi.org/10.18653/v1/N19-1162} {Cross-lingual alignment
  of contextual word embeddings, with applications to zero-shot dependency
  parsing}.
\newblock In \emph{Proceedings of the 2019 Conference of the North {A}merican
  Chapter of the Association for Computational Linguistics: Human Language
  Technologies, Volume 1 (Long and Short Papers)}, pages 1599--1613,
  Minneapolis, Minnesota. Association for Computational Linguistics.

\bibitem[{Schwenk et~al.(2019{\natexlab{a}})Schwenk, Chaudhary, Sun, Gong, and
  Guzm{\'{a}}n}]{Schwenk2019wikimatrix}
Holger Schwenk, Vishrav Chaudhary, Shuo Sun, Hongyu Gong, and Francisco
  Guzm{\'{a}}n. 2019{\natexlab{a}}.
\newblock \href {https://arxiv.org/abs/1907.05791} {{WikiMatrix: Mining 135M
  Parallel Sentences}}.
\newblock \emph{arXiv preprint arXiv:1907.05791}.

\bibitem[{Schwenk and Li(2018)}]{Schwenk2018}
Holger Schwenk and Xian Li. 2018.
\newblock \href {https://www.aclweb.org/anthology/L18-1560} {A corpus for
  multilingual document classification in eight languages}.
\newblock In \emph{Proceedings of the Eleventh International Conference on
  Language Resources and Evaluation ({LREC} 2018)}, Miyazaki, Japan. European
  Language Resources Association (ELRA).

\bibitem[{Schwenk et~al.(2019{\natexlab{b}})Schwenk, Wenzek, Edunov, Grave, and
  Joulin}]{Schwenk2019ccmatrix}
Holger Schwenk, Guillaume Wenzek, Sergey Edunov, Edouard Grave, and Armand
  Joulin. 2019{\natexlab{b}}.
\newblock \href {https://arxiv.org/abs/1911.04944} {{CCMatrix: Mining Billions
  of High-Quality Parallel Sentences on the WEB}}.
\newblock \emph{arXiv preprint arXiv:1911.04944}.

\bibitem[{Siddhant et~al.(2019)Siddhant, Johnson, Tsai, Arivazhagan, Riesa,
  Bapna, Firat, and Raman}]{Siddhant2019evaluating}
Aditya Siddhant, Melvin Johnson, Henry Tsai, Naveen Arivazhagan, Jason Riesa,
  Ankur Bapna, Orhan Firat, and Karthik Raman. 2019.
\newblock \href {https://arxiv.org/abs/1909.00437} {{Evaluating the
  Cross-Lingual Effectiveness of Massively Multilingual Neural Machine
  Translation}}.
\newblock \emph{arXiv preprint arXiv:1909.00437}.

\bibitem[{Smith et~al.(2017)Smith, Turban, Hamblin, and Hammerla}]{Smith2017}
Samuel~L. Smith, David H.~P. Turban, Steven Hamblin, and Nils~Y. Hammerla.
  2017.
\newblock \href {https://openreview.net/pdf?id=r1Aab85gg} {{Offline bilingual
  word vectors, orthogonal transformations and the inverted softmax}}.
\newblock In \emph{Proceedings of the 5th International Conference on Learning
  Representations (ICLR 2017)}.

\bibitem[{Snyder et~al.(2008)Snyder, Naseem, Eisenstein, and
  Barzilay}]{Snyder2008postagging}
Benjamin Snyder, Tahira Naseem, Jacob Eisenstein, and Regina Barzilay. 2008.
\newblock \href {https://www.aclweb.org/anthology/D08-1109} {Unsupervised
  multilingual learning for {POS} tagging}.
\newblock In \emph{Proceedings of the 2008 Conference on Empirical Methods in
  Natural Language Processing}, pages 1041--1050, Honolulu, Hawaii. Association
  for Computational Linguistics.

\bibitem[{S{\o}gaard et~al.(2018)S{\o}gaard, Ruder, and
  Vuli{\'c}}]{Sogaard2018}
Anders S{\o}gaard, Sebastian Ruder, and Ivan Vuli{\'c}. 2018.
\newblock \href {https://doi.org/10.18653/v1/P18-1072} {On the limitations of
  unsupervised bilingual dictionary induction}.
\newblock In \emph{Proceedings of the 56th Annual Meeting of the Association
  for Computational Linguistics (Volume 1: Long Papers)}, pages 778--788,
  Melbourne, Australia. Association for Computational Linguistics.

\bibitem[{Song et~al.(2019)Song, Tan, Qin, Lu, and Liu}]{Song2019mass}
Kaitao Song, Xu~Tan, Tao Qin, Jianfeng Lu, and Tie-Yan Liu. 2019.
\newblock \href {http://proceedings.mlr.press/v97/song19d.html} {{MASS}: Masked
  sequence to sequence pre-training for language generation}.
\newblock In \emph{Proceedings of the 36th International Conference on Machine
  Learning}, volume~97, pages 5926--5936, Long Beach, California, USA. PMLR.

\bibitem[{Stojanovski et~al.(2019)Stojanovski, Hangya, Huck, and
  Fraser}]{stojanovski2019lmu}
Dario Stojanovski, Viktor Hangya, Matthias Huck, and Alexander Fraser. 2019.
\newblock \href {https://doi.org/10.18653/v1/W19-5344} {The {LMU} munich
  unsupervised machine translation system for {WMT}19}.
\newblock In \emph{Proceedings of the Fourth Conference on Machine Translation
  (Volume 2: Shared Task Papers, Day 1)}, pages 393--399, Florence, Italy.
  Association for Computational Linguistics.

\bibitem[{Vania and Lopez(2017)}]{Vania2017}
Clara Vania and Adam Lopez. 2017.
\newblock \href {https://doi.org/10.18653/v1/P17-1184} {From characters to
  words to in between: Do we capture morphology?}
\newblock In \emph{Proceedings of the 55th Annual Meeting of the Association
  for Computational Linguistics (Volume 1: Long Papers)}, pages 2016--2027,
  Vancouver, Canada. Association for Computational Linguistics.

\bibitem[{Vuli{\'c} et~al.(2019)Vuli{\'c}, Glava{\v{s}}, Reichart, and
  Korhonen}]{Vulic2019}
Ivan Vuli{\'c}, Goran Glava{\v{s}}, Roi Reichart, and Anna Korhonen. 2019.
\newblock \href {https://doi.org/10.18653/v1/D19-1449} {Do we really need fully
  unsupervised cross-lingual embeddings?}
\newblock In \emph{Proceedings of the 2019 Conference on Empirical Methods in
  Natural Language Processing and the 9th International Joint Conference on
  Natural Language Processing (EMNLP-IJCNLP)}, pages 4407--4418, Hong Kong,
  China. Association for Computational Linguistics.

\bibitem[{Vuli{\'c} and Korhonen(2016)}]{Vulic2016seedlexicon}
Ivan Vuli{\'c} and Anna Korhonen. 2016.
\newblock \href {https://doi.org/10.18653/v1/P16-1024} {On the role of seed
  lexicons in learning bilingual word embeddings}.
\newblock In \emph{Proceedings of the 54th Annual Meeting of the Association
  for Computational Linguistics (Volume 1: Long Papers)}, pages 247--257,
  Berlin, Germany. Association for Computational Linguistics.

\bibitem[{Wang et~al.(2018)Wang, Singh, Michael, Hill, Levy, and
  Bowman}]{Wang2019glue}
Alex Wang, Amanpreet Singh, Julian Michael, Felix Hill, Omer Levy, and Samuel
  Bowman. 2018.
\newblock \href {https://doi.org/10.18653/v1/W18-5446} {{GLUE}: A multi-task
  benchmark and analysis platform for natural language understanding}.
\newblock In \emph{Proceedings of the 2018 {EMNLP} Workshop {B}lackbox{NLP}:
  Analyzing and Interpreting Neural Networks for {NLP}}, pages 353--355,
  Brussels, Belgium. Association for Computational Linguistics.

\bibitem[{Wu et~al.(2019)Wu, Conneau, Li, Zettlemoyer, and
  Stoyanov}]{wu2019emerging}
Shijie Wu, Alexis Conneau, Haoran Li, Luke Zettlemoyer, and Veselin Stoyanov.
  2019.
\newblock \href {https://arxiv.org/abs/1911.01464} {Emerging cross-lingual
  structure in pretrained language models}.
\newblock \emph{arXiv preprint arXiv:1911.01464}.

\bibitem[{Wu and Dredze(2019)}]{Wu2019}
Shijie Wu and Mark Dredze. 2019.
\newblock \href {https://doi.org/10.18653/v1/D19-1077} {Beto, bentz, becas: The
  surprising cross-lingual effectiveness of {BERT}}.
\newblock In \emph{Proceedings of the 2019 Conference on Empirical Methods in
  Natural Language Processing and the 9th International Joint Conference on
  Natural Language Processing (EMNLP-IJCNLP)}, pages 833--844, Hong Kong,
  China. Association for Computational Linguistics.

\bibitem[{Xu et~al.(2018)Xu, Yang, Otani, and Wu}]{xu2018unsupervised}
Ruochen Xu, Yiming Yang, Naoki Otani, and Yuexin Wu. 2018.
\newblock \href {https://doi.org/10.18653/v1/D18-1268} {Unsupervised
  cross-lingual transfer of word embedding spaces}.
\newblock In \emph{Proceedings of the 2018 Conference on Empirical Methods in
  Natural Language Processing}, pages 2465--2474, Brussels, Belgium.
  Association for Computational Linguistics.

\bibitem[{Zhang et~al.(2017{\natexlab{a}})Zhang, Liu, Luan, and
  Sun}]{zhang2017adversarial}
Meng Zhang, Yang Liu, Huanbo Luan, and Maosong Sun. 2017{\natexlab{a}}.
\newblock \href {https://doi.org/10.18653/v1/P17-1179} {Adversarial training
  for unsupervised bilingual lexicon induction}.
\newblock In \emph{Proceedings of the 55th Annual Meeting of the Association
  for Computational Linguistics (Volume 1: Long Papers)}, pages 1959--1970,
  Vancouver, Canada. Association for Computational Linguistics.

\bibitem[{Zhang et~al.(2017{\natexlab{b}})Zhang, Liu, Luan, and
  Sun}]{zhang2017earth}
Meng Zhang, Yang Liu, Huanbo Luan, and Maosong Sun. 2017{\natexlab{b}}.
\newblock \href {https://www.aclweb.org/anthology/D17-1207} {Earth mover's
  distance minimization for unsupervised bilingual lexicon induction}.
\newblock In \emph{Proceedings of the 2017 Conference on Empirical Methods in
  Natural Language Processing}, pages 1934--1945, Copenhagen, Denmark.
  Association for Computational Linguistics.

\end{thebibliography}
\bibliographystyle{acl_natbib}

\end{document}